\pdfoutput=1

\documentclass[11pt]{article}

\usepackage{acl}
\usepackage[greek,english]{babel}

\usepackage{times}
\usepackage{latexsym}
\usepackage{amsfonts}
\usepackage{comment}
\usepackage{booktabs}
\usepackage{graphicx}
\usepackage{multirow}
\usepackage{longtable}
\usepackage{algorithm}
\usepackage{amsmath}
\usepackage[T1]{fontenc}
\usepackage{cleveref}
\crefname{section}{§}{§§}

\usepackage[utf8x]{inputenc}

\usepackage{microtype}
\usepackage{xcolor}
\usepackage{cleveref}
\crefname{section}{§}{§§}
\definecolor{c1}{HTML}{4e79a7}%
\definecolor{c2}{HTML}{f28e2b}%
\definecolor{c3}{HTML}{009E73}%
\definecolor{c4}{HTML}{56B4E9}%
\definecolor{c5}{HTML}{CC79A7}%
\definecolor{c6}{HTML}{E69F00}%
\definecolor{c7}{HTML}{844E4D}%
\definecolor{c8}{HTML}{2D512A}%

\definecolor{oorange}{HTML}{d95f02}
\definecolor{bblue}{HTML}{7570b3}
\definecolor{ggreen}{HTML}{1b9e77}
\definecolor{ppurple}{HTML}{e37fbb}
\definecolor{lgreen}{HTML}{9CD24A}
\definecolor{yyellow}{HTML}{FFD52D}
\definecolor{ggold}{HTML}{E1BC89}
\definecolor{ggray}{HTML}{AAAAAA}

\title{Computational Discovery of Chiasmus in Ancient Religious Text}

\author{
    {\bf Hope McGovern}\textsuperscript{1} ~~
    {\bf Hale Sirin}\textsuperscript{2} ~~
    {\bf Tom Lippincott}\textsuperscript{2} ~~ \\
    \textsuperscript{1} Department of Computer Science \& Technology, University of Cambridge, U.K. \\
    \textsuperscript{2} Center for Digital Humanities, Johns Hopkins University, Baltimore, U.S.A. \\
    \small \textsuperscript{1} \texttt{hope.mcgovern@cl.cam.ac.uk}
    \hspace{3mm}
    \small \textsuperscript{2} \texttt{\{hsirin1, tom.lippincott\}@jhu.edu}
    \hspace{3mm} \\
}

\begin{document}
\maketitle
\begin{abstract}
Chiasmus, a debated literary device in Biblical texts, has captivated mystics while sparking ongoing scholarly discussion. In this paper, we introduce the first computational approach to systematically detect chiasmus within Biblical passages. Our method leverages neural embeddings to capture lexical and semantic patterns associated with chiasmus, applied at multiple levels of textual granularity (half-verses, verses). We also involve expert annotators to review a subset of the detected patterns. Despite its computational efficiency, our method achieves robust results, with high inter-annotator agreement and system precision@$k$ of 0.80 at the verse level and 0.60 at the half-verse level. We further provide a qualitative analysis of the distribution of detected chiasmi, along with selected examples that highlight the effectiveness of our approach.\footnote{All code and data available at \url{https://github.com/comp-int-hum/literary-translation}}

\end{abstract}

\section{Introduction}
Chiasmus is a topic which fascinates Bible scholars. Most simply and broadly understood, \textit{chiasmus}, or chiasm, denotes a sequence of textual units that intentionally exhibit a semantic or poetic symmetry. A clear chiastic example in English is JFK's adage (with corresponding textual units in the same color):
\begin{quote}
\small
    Ask not what \textcolor{blue}{your country} \textcolor{red}{can do} \textcolor{orange}{for you,} 
    
    but what \textcolor{orange}{you} \textcolor{red}{can do} \textcolor{blue}{for your country.}
\end{quote} 
The name derives from the Greek letter $\chi$, `chi', which looks like an English `X' and is used to illustrate the structure of a chiasmus: e.g. ABB'A', as shown in \autoref{tab:chiasm_ex}. Chiasmi may be even or odd (i.e. having an unpaired distinct center), and may have an arbitrary number of lines.

While chiasmus in English is associated with high oratory skill \cite{bothwell-etal-2023-introducing}, it is exceedingly rare as a rhetorical device in modern language:
English experts trawling through a corpus of Winston Churchill's works found only seven chiasmi out of a total of $\sim$200 speeches \cite{dubremetz-nivre-2015-rhetorical}.  
However, chiasmus is extremely common in ancient literature and oratory 
\cite{Welch1981ChiasmusIA}. It has been known to be a common rhetorical feature of Ancient Hebrew poetry since the 1740s \cite{lowth1839lectures}.

\begin{table}[h]
\footnotesize
\centering
\begin{tabular}{lp{6cm}}
\toprule
A  & [...] Let them be turned back and disappointed \textcolor{blue}{who devise evil against me!}  \\
\multirow{2}{*}{\quad B}  &  \quad \textcolor{red}{Let them be like chaff before the wind,} \textcolor{ggreen}{with} \\
& \quad \textcolor{ggreen}{the angel of the LORD driving them away!}  \\
\multirow{2}{*}{\quad B'} &  \quad \textcolor{red}{Let their way be dark and slippery,} \textcolor{ggreen}{with} \\
& \quad \textcolor{ggreen}{the angel of the LORD pursuing them!} \\
A' & For without cause \textcolor{blue}{they hid their net for me; without cause they dug a pit for my life.}\\
\bottomrule
\end{tabular}
\caption{\textbf{The `X' pattern of chiasm in Psalm 35:4-7 (ESV).} Pairs (A, A') and (B, B') exhibit repeated phrases and conceptual links.}
\label{tab:chiasm_ex}
\end{table}
While most scholars agree that chiasmus is a facet of Ancient Near Eastern writings, there is much debate about its prevalence, purpose, and location. 
Biblical scholars have proposed its use to underscore characterization in narrative passages \cite{chiasmus_characterization}, as a poetic device in the Psalms \cite{Martin2018Psalms}, and to capture ritualistic language in legal documents \cite{chiasmus_structural}. However, a lack of quantitative methods for Biblical chiasmus renders the task of detection a laborious and subjective one. We provide a straightforward method to computationally formalize and detect chiasmi.





Unlike previous work which utilized handcrafted features and a log-linear model to detect fine-grained instances of chiasmus in English prose \cite{dubremetz-nivre-2017-machine}, we use a statistical method based on cosine distance from line-level embedded representations of text. The use of embeddings instead of only lemmata allows us to include semantic information between lines that form a chiastic structure, enabling a more nuanced definition of chiasmus in line with rhetorical intention. This approach is supported by recent work in rhetorical device detection \cite{schneider}, and includes the repetition of words, phrases, grammatical structures, or (identical or antithetical) concepts as part of the chiastic parallels. In contrast with \citet{schneider}, our method is extensible to various sizes of chiasmus; that is, those of just four lines long or of 100 lines long, and is language-agnostic, whereas previous work has focused only on fine-grained, intra-line chiasmus in English or German. In this study, we analyze both half-verses and verses as units so that a chiasmus within the same verse can also be captured (i.e. ``The Sabbath was made for man, not man for the Sabbath''). We formalize the notion of Biblical chiasmus thoroughly in \cref{sec:formalism_chiasm}.

Our main contributions are as follows:
\begin{enumerate}
    \item We show that multilingual embedding spaces may be effectively used to detect rhetorical phenomena such as chiasmus in ancient manuscripts.
    \item We provide, for the first time, a mathematical formalism of Biblical chiasmus and provide a computational algorithm for its detection.
    \item Our method is computationally efficient and achieves robust results, with high inter-annotator agreement and system precision@$k$ of 0.80 at the verse level and 0.60 at the half-verse level. 
    \item We contribute to Classics and Biblical Studies by providing a qualitative analysis of the distribution of detected chiasmi, along with selected examples that highlight the effectiveness of our approach. 

\end{enumerate}

\section{Method}

\subsection{Data}
We use as our primary source the Translator’s Amalgamated Hebrew Old Testament (TAHOT)\footnote{\url{www.STEPBible.org}}, which is based on the Leningrad Codex -- the oldest complete extant version of the Hebrew Old Testament. Note that modern English translations follow a versification system that is at times slightly different to the Hebrew text due to a difference in textual traditions. We use the Hebrew versification system to better uncover chiasmi as they may be in the original text. \texttt{N.B.} We carry out all detection experiments using the Hebrew text, but for clarity and accessibility, report English translations\footnote{We release a formatted version of STEP Bible's data, including translations, on the Huggingface Hub. DOI: \href{https://huggingface.co/datasets/hmcgovern/original-language-bibles-hebrew}{10.57967/hf/4174}.} in tables and figures.

We segment the text into two levels: verses and half-verses. In the Hebrew text, half-verses are naturally marked by the cantillation symbol, \textit{atnach}, which typically separates the two halves of a verse. We consider up to and including the word with the \textit{atnach} to be the first half, while the remainder is the second half. We then remove all vocalizations and cantillation symbols before embedding. 

\subsection{Formalizing Chiasmus}
\label{sec:formalism_chiasm}

The first step in our method involves constructing a cosine similarity matrix, denoted as \( S \), based on feature vectors extracted from the text via E5, a multilingual embedding model \cite{e52024multilingual}\footnote{We use the `small' variant of this model, with 118M parameters.}. Our method is similar to that of \citet{burns2021profiling}, which uses pairwise cosine similarity of embedded representations to identify intertextual phrases in Latin.

Each element \( S_{ij} \) represents the cosine similarity between the feature vectors of textual units \( i \) and \( j \). Next, we identify potential chiastic structures by focusing on matching groups of text pairs, such as \( A \) and \( A' \), \( B \) and \( B' \), and so forth. For each potential chiastic structure, we compute the \textit{chiasmus score} \( \mu_{\text{chiasmus}} \), which is the average cosine similarity of these matching pairs:

\begin{equation}
\mu_{\text{chiasmus}} = \frac{1}{k} \sum_{i=1}^{k} S_{\text{pair}(i)}
\end{equation}

where \( \text{pair}(i) \) refers to the indices of the matching pairs (e.g., \( A \) and \( A' \), \( B \) and \( B' \)). To assess the distinctiveness of this chiastic structure, we compute the average of all non-pair similarities, denoted \( \mu_{\text{non-pair}} \), which includes comparisons such as \( S_{A,B}, S_{B,C'} \), and others:

\begin{equation}
\mu_{\text{non-pair}} = \frac{1}{n} \sum_{i,j \in \text{non-pair}} S_{ij}
\end{equation}

Our final score for each window is computed as the difference between these two averages:

\begin{equation}
\text{Final Score} = \mu_{\text{chiasmus}} - \mu_{\text{non-pair}}
\end{equation}

To detect chiasmi across the text, we apply this method in a \textit{sliding window} fashion, where each starting position in the text serves as a potential beginning of a chiastic structure. The length of the sliding window, $N$, is fixed for each experiment, and we test several different $N$ values, analyzing the aggregated results. We ensure that chiasmi do not cross book boundaries by disallowing matches across these divisions.

Finally, we standardize the chiasmus scores across the text by calculating their z-scores. The z-score \( z_i \) for each window \( i \) is determined by:

\begin{equation}
z_i = \frac{\mu_{\text{chiasmus},i} - \mu_{\text{chiasmus},\text{mean}}}{\sigma_{\text{chiasmus}}}
\end{equation}

where \( \mu_{\text{chiasmus},\text{mean}} \) and \( \sigma_{\text{chiasmus}} \) are the mean and standard deviation of all chiasmus scores, respectively. We classify chiastic structures as significant if their z-scores exceed a threshold of three (3) standard deviations above the mean, thereby identifying statistically salient chiasmi within the text.

\subsection{Why not use an LLM?}
While large language models (LLMs) have enabled remarkable advances in a wide variety of NLP tasks, data contamination concerns and a lack of explainability limit their scope of usefulness for chiasmus detection in Biblical text. 

Preliminary exploration revealed that some LLMs have a propensity to generate verbatim copyrighted English translations (e.g., the ESV) from Ancient Hebrew source passages. This behavior suggests that the extensive availability of online Biblical commentaries, which may reference chiastic structure, is likely included in web-based training corpora. Consequently, the outputs of LLMs risk being skewed by prior exposure to human annotation \cite{BalloccuLeakCheatRepeat}. 

Furthermore, the lack of transparency in LLM-generated responses poses a significant barrier for adoption in scholarly contexts. Biblical scholars, who are our primary target audience, require interpretable and verifiable methods rather than opaque, black-box solutions. Additionally, our aim extends beyond merely detecting chiasmi; we seek to formalize the concept mathematically, thereby offering a rigorous and standardized framework for discussing what remains a somewhat ambiguous topic. Such a formalism could serve as a valuable tool for facilitating scholarly discourse and advancing the study of chiasmus.

\section{Experiments}
We run our model over the Hebrew Old Testament, considering every line or half-line as a potential starting position and length $(N)$ of chiasmus to be in the range of four to eight ($N \in [4,8]$). We take the top-50 highest-scoring outputs for both half-verse and verse grouping and evaluate them via human annotation. Annotation guidelines and results are found in \cref{sec:human_annotation}. We use top-$k$ precision as our evaluation metric as we are primarily interested in creating a tool for scholars to find the most-promising candidates for chiasmus to further examine. 

\autoref{tab:stats} presents an overview of the system’s output for chiastic structures at the half-verse and verse levels. A total of 1,896 chiastic structures were identified at the half-verse level, with an average length of 5.93 textual units ($\pm$1.34) and an average score of 0.32 ($\pm$0.1). For verse-level groupings, 879 chiastic structures were found, with an average length of 6.01 lines ($\pm$1.38) and an average score of 0.29 ($\pm$0.08). The book of Genesis contains the highest number of half-verse chiasmi, while Numbers contains the most verse-level chiasmi.

\begin{table}[]
\footnotesize
\begin{tabular}{@{}lllll@{}}
\toprule

                                    &                 & \textbf{Half-Verse}     & \textbf{Verse}           &  \\ \midrule
\multirow{4}{*}{\rotatebox[origin=c]{90}{\tiny Full Output }} & Num. Found      & 1896           & 879             &  \\
                                    & Top Book        & Genesis        & Numbers         &  \\
                                    & Avg. Length      & 5.93  $\pm$ 1.34         & 6.01  $\pm$ 1.38          &  \\
                                    & Avg. Score       & 0.32 $\pm$ 0.1 & 0.29 $\pm$ 0.08 &  \\ \cmidrule(lr){2-4}
\multirow{2}{*}{\rotatebox[origin=c]{90}{\tiny Annotated}}   & System Precision@$k$ &      0.60          & 0.80            &  \\
                                    & Cohen Kappa ($\kappa$)     &    0.76            & 0.89            & \\ 
                                    & Top Genre & Narrative & Narrative \\\bottomrule
\end{tabular}
\caption{\textbf{Summary of detected chiasmi.} 2700+ chiasmi were detected at the verse and half-verse level. The highest number of chiasmi was found in the Book of Genesis and Book of Numbers. Both the precision and the inter-annotator agreement increase for the verse-level chiasmi.}
\label{tab:stats}
\end{table}
\begin{figure}[t]
    \centering
    \includegraphics[width=1.1\linewidth]{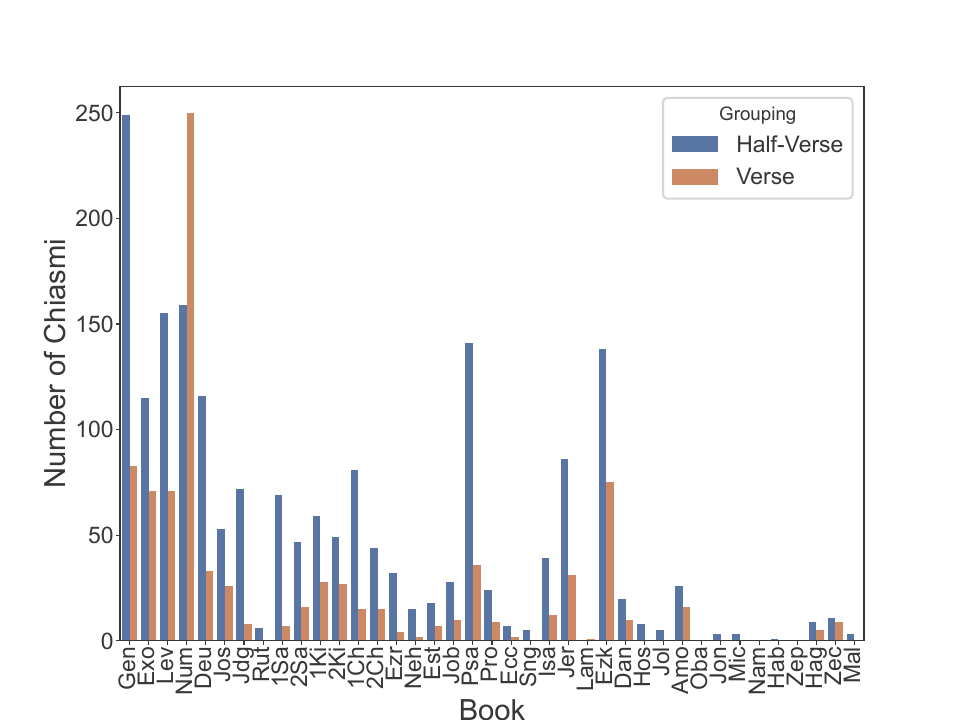}
    \caption{\textbf{Total number of chiasmi per Book at verse and half-verse level.} While some books tend to have more chiasmi overall, this figure shows whether verse-level or half-verse-level chiasmi are more prevalent in each book.}
    \label{fig:counts}
\end{figure}

As shown in \autoref{fig:counts}, the number of detected chiastic structures varies across books of the Bible, with more instances found at the half-verse level than at the verse level for all books. Notably, certain books exhibit disproportionately higher numbers of half-verse chiasmi, particularly Genesis, 1 Samuel, Judges, 1 Chronicles, Psalms, Jeremiah, and Ezekiel. This trend is consistent with the literary nature of these texts: Psalms, Jeremiah, and Ezekiel include significant poetic sections, where half-verse chiastic structures are more prominent, while Genesis features dense narrative and highly literary passages as well as many formulaic genealogies. The high counts in 1 Samuel, Judges, and 1 Chronicles, which are historical books, likely reflect the system’s identification of formulaic narrative patterns, such as the recurring descriptions of the kingly line of Israel (e.g., ``X became king, reigned for Y years, and did evil in the sight of the Lord'').

\begin{table*}[h]
\begin{tabular}{lp{15cm}}
\toprule
A & \textcolor{blue}{And there was evening and there was morning, the fourth day.} \\
\multirow{2}{*}{\quad B} & \quad \textcolor{red}{And God said, ``Let the waters swarm with swarms of living creatures}, and \textcolor{ggreen}{let birds fly above the} \\
\quad & \quad \textcolor{ggreen}{earth} across the expanse of the heavens.'' \\
\multirow{3}{*}{\quad \quad C} & \quad \quad So God created the great sea creatures and every living creature that moves, with which the \\
 & \quad \quad waters swarm, according to their kinds, and every winged bird according to its kind. And God \\
  & \quad \quad saw that it was good.   \\
\multirow{2}{*}{\quad B'} & \quad And God blessed them, saying, \textcolor{red}{“Be fruitful and multiply and fill the waters in the seas}, and \\
& \quad \textcolor{ggreen}{let birds multiply on the earth.}”  \\
A' & \textcolor{blue}{And there was evening and there was morning, the fifth day.}
\\ \bottomrule
\end{tabular}
\caption{\textbf{English translation of a positive example of a chiasmus automatically detected by our method.} Gen 1:19-23 (ESV) }
\label{tab:gen_1}
\end{table*}
\subsection{Human Annotation}
\label{sec:human_annotation}
The top-50 scoring half-verse and verse chiasmi were manually reviewed by the first two authors, who both have graduate-level training in ancient languages and literature\footnote{While the chiasmus identification is done entirely in Hebrew, the annotators use a literal English translation following Hebrew word order alongside the Hebrew text for easier inspection.}. Given a three-class rubric, they were asked to determine whether the set of verses of half-verses identified by the model exhibited (1) \textit{chiastic repetition}: a chiastic structure of repetition formed either through lexical or semantic textual units, (2) \textit{non-chiastic repetition}: lexical or semantic repetition of textual units, but not in a discernibly chiastic way, or (3) \textit{no repetition}: no discernible parallel or repeating content. Cohen's Kappa ($\kappa$), used to quantify inter-annotator agreement, is 0.76 and 0.89 for half-verses and verses, respectively, indicating strong agreement between the annotators. 

Two verse-level passages and four half-verse level passages were putative between chiastic repetition and non-chiastic repetition, while there were only two (both half-verse) passages that were disputed between no repetition and chiastic repetition. In other words, annotators were nearly always in agreement over which passages had elements of structural repetition, but discerning between chiastic and non-chiastic repetition poses a slightly more difficult challenge. 

Considering ``true'' chiasmi to be those marked as chiastic by both annotators, we achieve a system precision@$k$ of \textbf{0.60} for half-verses and \textbf{0.80} for verses.  In both experiments, the majority of top-scoring chiasmi are found in narrative sections of text.

Interestingly, passages classified as \textit{non-chiastic repetition} often involved formulaic or ritualistic language, which could be of interest to scholars seeking computational methods for identifying such patterns in texts. We find 29 examples of this across the top 100 collectively. Only 3 of the top 100, or 3\%, of the top-scoring passages belonged to the \textit{no repetition} class. 

\section{Discussion}
Several qualitatively interesting examples of chiasmus were identified by our method, highlighting the richness of the Biblical texts and the alignment with existing literary scholarship. One notable example is Genesis 1:19-23, as shown in \autoref{tab:gen_1}. This five-line chiasmus, positively identified by both annotators, exhibits clear lexical parallels between its paired sections. The chiastic structure here emphasizes the order and the rhetorical intentionality in the Creation narrative, underscoring God's repeated affirmation that His creation is ``good''. This example aligns with scholarly interpretations that highlight the poetic nature of the Creation account.

Other significant examples include the story of Jacob stealing Esau's birthright, where the chiastic structure reflects the tension and reversal of fortune between the brothers. Similarly, the account of Isaac and Abraham and the sacrificial lamb contains a chiasmus that heightens the dramatic and theological impact of the narrative, as God intervenes at the critical moment.

The method also uncovered a clear chiasmus in God’s covenant with Noah after the flood, where the repetitive structure emphasizes God's promise of restoration and the symbolic importance of the `bow' in the clouds. Additionally, in Ezekiel’s poetic description of the image of the glory of the LORD, chiastic elements serve to enhance the vividness and majesty of the vision, a hallmark of Ezekiel's prophetic style. Illustrations of these chiasmi may be seen in \cref{appendix:chiasm_case_study}.

Notably, many instances of God's reported speech are presented in chiastic or poetic form, which may suggest an intentional literary quality meant to convey authority and solemnity. These findings further support the hypothesis that chiasmus is often employed for rhetorical and theological purposes in Biblical texts.

\section{Conclusion}
Our approach demonstrates the ability to uncover intricate literary patterns that might otherwise be overlooked, providing valuable insights for scholars of Biblical texts, political oratory, and literary studies. This example, along with our overall findings, underscores the importance of advanced computational techniques in literary analysis and supports the broader application of our method for discovering chiasmi across various texts and translations. One future step is using a chiasmus detection method to create a labeled corpus of chiasmi within the Bible, particularly the Psalms, for scholarly exploration.




\section*{Limitations}
In this study, we only investigate chiastic structures at the verse-level and half-verse-level. However, chiasmi can also be identified at the narrative level, where narrative segments \textit{topically} form a chiastic plot structure, such as the narrative of the flood in Genesis. We exclude this type since it exhibits many fewer lexical features and is overall less precisely defined in the scholarship.

\section*{Acknowledgments}
We would like to thank scholars at Tyndale House, especially Ellie Wiener and Caleb J. Howard, for useful discussions about Biblical Hebrew. Thanks to Andrew Caines for his helpful comments on previous drafts.
Hope McGovern's work is supported by the Woolf Institute for Interfaith Relations and the Cambridge Trust

\bibliography{literary_translation}
\bibliographystyle{acl_natbib}
\onecolumn

\appendix

\section{Chiasms Referenced in the Discussion Section}
\label{appendix:chiasm_case_study}

\begin{table*}[h]
\begin{tabular}{lp{12cm}}
\toprule
A &  And Esau said to Jacob, ``Let me eat some of that red stew, for I am exhausted!'' (Therefore his name was called Edom.)  \\
\multirow{2}{*}\quad{B} & \quad  Jacob said, ``Sell me your birthright now.''\\
{\quad \quad C} & \quad \quad Esau said, ``I am about to die; of what use is a birthright to me?''\\

{\quad B'} & \quad Jacob said, ``Swear to me now.'' So he swore to him and sold his birthright to Jacob. \\
A' & Then Jacob gave Esau bread and lentil stew, and he ate and drank and rose and went his way. Thus Esau despised his birthright.\\ \bottomrule
\end{tabular}

\caption{\textbf{The story of Jacob stealing Esau's birthright.} English translation of a chiasmus automatically detected by our method. Gen 25:30-34 (ESV) }
\label{tab:esau}
\end{table*}

\begin{table*}[h]
\begin{tabular}{lp{13cm}}
\toprule
A  & And God said, ``This is the sign of the covenant that I make between me and you and every living creature that is with you, for all future generations:  \\
{\quad B}  &  \quad I have set my bow in the cloud, and it shall be a sign of the covenant \\ & \quad between me and the earth.  \\
{\quad B'} &  \quad When I bring clouds over the earth and the bow is seen in the clouds, \\
A' & I will remember my covenant that is between me and you and every living creature of all flesh. And the waters shall never again become a flood to destroy all flesh. \\
\bottomrule
\end{tabular}
\caption{\textbf{God's covenant with Noah after the flood.} English translation of a chiasmus automatically detected by our method. Gen 9:12-15 (ESV)}
\label{tab:noah}
\end{table*}




\end{document}